%% file: main.tex
\documentclass[preprint,review,1p,12pt,times]{elsarticle}
\usepackage{amssymb}
\usepackage{booktabs}

\usepackage{amsmath}
\usepackage{algorithm}
\usepackage{algorithmic}
\usepackage{xcolor}
\usepackage[table]{xcolor}  
\usepackage{booktabs}

\journal{Pattern Recognition}

\begin{document}

\begin{frontmatter}

\title{Beyond Consistency: Preserving Temporal Structure in Zero-Shot Video Editing} 

\author[label1,label2]{Deyin Liu}
\author[label1]{Yisheng Ding}
\author[label1]{Zhe Jin}
\author[label2]{Xiatian Zhu}
\author[label2]{Anjan Dutta}
\author[label3]{Lin Wu}
\affiliation[label1]{organization={Anhui University},
            addressline={No 111, Jiulong Road},
            city={Hefei},
            postcode={230601},
            state={Anhui Province},
            country={China}}

\affiliation[label2]{organization={Surrey Institute for People-Centred Artificial Intelligence, University of Surrey},
            city={Guildford},
            postcode={GU2 7XH},
            state={Surrey},
            country={UK}}

\affiliation[label3]{organization={University of Warwick},
            addressline={Gibbet Hill Road}, 
            city={Coventry},
            postcode={CV4 7AL}, 
            country={UK}}

\begin{abstract}
Existing zero-shot video editing methods rely on pre-trained diffusion models, successfully achieving spatial control and basic temporal consistency but fundamentally fail to preserve the video's original temporal structure. This distinction is critical: temporal consistency ensures visual smoothness, but temporal structure dictates the video's high-level narrative, rhythm, and semantic flow. Without this preservation, the edited output, especially for long videos with complex semantic variations, becomes narratively incoherent and semantically ambiguous. 
To address this limitation, we introduce a novel zero-shot editing approach that, for the first time, explicitly focuses on preserving the source video’s temporal structure. We achieve this by adaptively partitioning the video into semantically distinct clips based on feature similarity and selecting a representative anchor frame for each clip. To enhance both intra-clip fidelity and computational efficiency, we design a clip-adaptive token merging strategy which leverages the anchor's semantic dominance to stabilize the editing. Furthermore, we employ an alternating combination strategy that ensures seamless inter-clip transitions while maintaining semantic distinction. Extensive experiments demonstrate that our method achieves state-of-the-art results, successfully balancing the preservation of original temporal structure with computational efficiency, and setting a new benchmark for zero-shot video editing fidelity.

\end{abstract}


\end{frontmatter}

\section{Introduction}
\label{sec1}

Diffusion models, which were originally inspired from thermodynamics \citep{DBLP:conf/icml/Sohl-DicksteinW15} and then advanced by  DDPM \citep{DBLP:conf/nips/HoJA20, DBLP:conf/icml/NicholD21} and Latent Diffusion Models \citep{DBLP:conf/cvpr/RombachBLEO22,ZHEN2025111479}, have shown significant success in image synthesis \citep{DBLP:conf/nips/DhariwalN21}  and editing \citep{CHEN2026112583,LIU2026112510,XIAO2023109458} tasks, laying a solid foundation for visual content creation \citep{IANCHAN2026112614,DBLP:journals/pami/CroitoruHIS23}.Naturally, researchers desire to translate these advances into the video domain.

Although some large-scale trained Text-2-Video (T2V) diffusion models \citep{VDM,Imagen-video} have emerged, to reduce the memory and computation overhead and ensure flexibility, considerable existing methods adapt pre-trained Text-2-Image (T2I) diffusion models \citep{DBLP:conf/cvpr/RombachBLEO22,Imagen} for video editing tasks. Among them, one-shot tuning approaches represented by Tune-A-Video \citep{DBLP:conf/iccv/WuGWLGSHSQS23} and Video-P2P \citep{DBLP:conf/cvpr/LiuZ00J24} capitalize on tuning on pre-trained T2I models for one-shot video editing, which however still involve large
computation derived from fine-tuning or optimization \citep{Towards}.

\begin{figure*}[t]
    \begin{center}
        \includegraphics[width=1.0\linewidth]{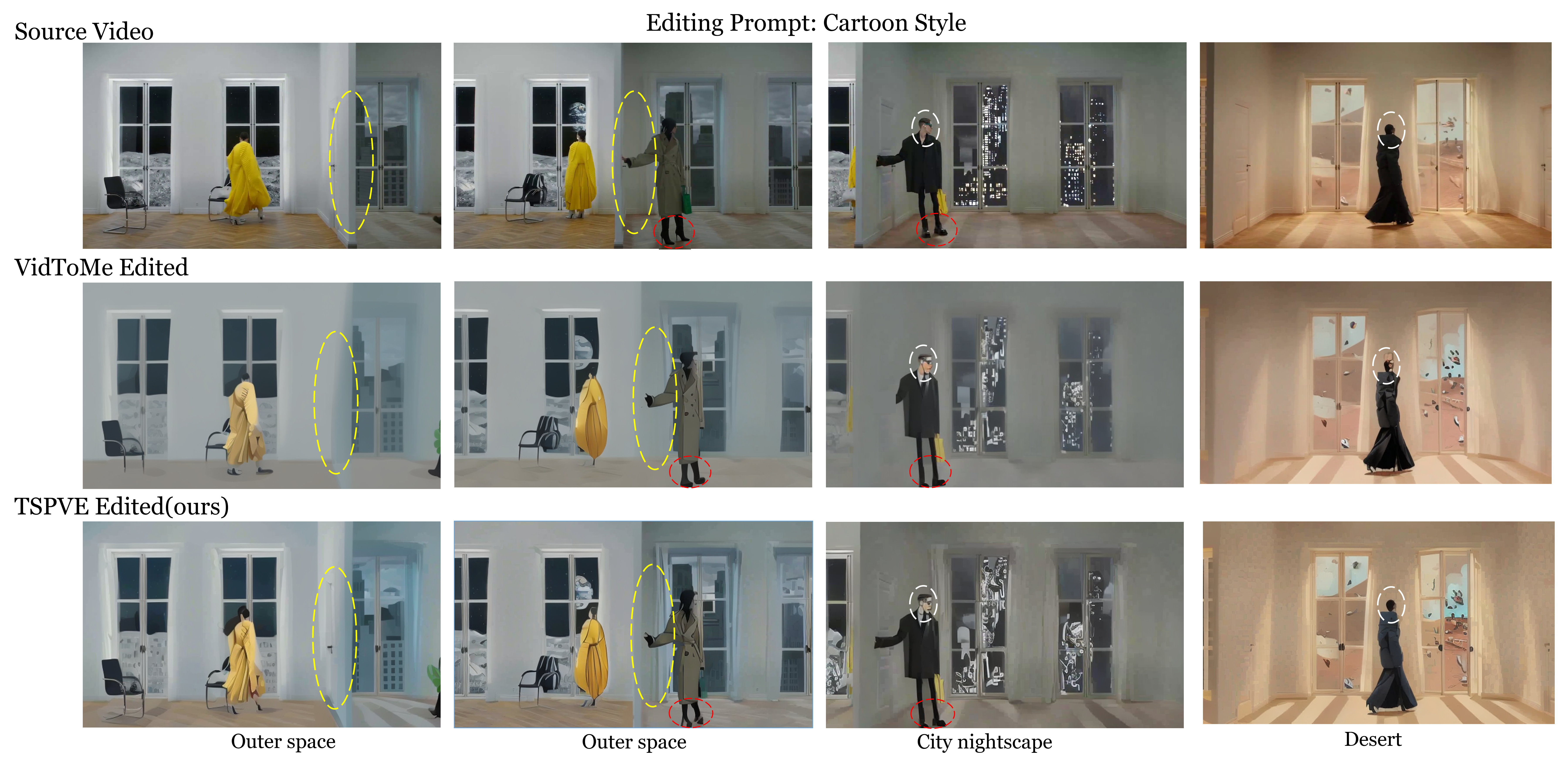}
        \caption{
Qualitative comparison with failure cases of one prior representative method. \textbf{Top row:} Sampled frames from a long source video sequence which describes some Illusory scene transformations that a model successively crossing multiple rooms while keeping changing its identities and outfits, with each room corresponding to a semantically independent scene. \textbf{Middle row:} Guided by the editing prompt "Cartoon style" to change the video's style, the existing method VidToMe only forces global temporal consistency, causing cross-scene distortions such as the smoothed structure of wall (yellow ellipses), the same shoe shape of different models (red ellipses), and the similar textures and colors of different heads (white ellipses). \textbf{Bottom row:} Our proposed TSPVE allows the edited frames to retain temporal semantic variations in the source video, preserving the temporal structure accurately. (Zoom in, better for observation).
}
        \label{fig:framework-related}
    \end{center} 
\end{figure*}

Recently, zero-shot video editing paradigm leverages the popular pre-trained T2I diffusion models (e.g., Stable Diffusion \citep{DBLP:conf/cvpr/RombachBLEO22}) to modify visual content of a video to align with the target edit prompt while retaining the original information such as spatial structure and motion \citep{DBLP:conf/iccv/QiCZLWSC23,DBLP:conf/iccv/CeylanHM23}. 
Relying on the controlling strategies of existing image editing methods \citep{DBLP:conf/cvpr/TumanyanGBD23,DBLP:conf/iccv/ZhangRA23} to preserve the visual appearance or spatial structure of the source video in generative frames, many zero-shot video editing methods focus mainly on how to maintain temporal consistency throughout the edited video. For example, two representative methods manipulate the multi-frame visual tokens from the diffusion models, by either token propagation based on inter-frame correspondence \citep{DBLP:conf/iclr/GeyerBBD24}  or token merging and unmerging \citep{DBLP:conf/cvpr/LiMYY24} during the denoising process, achieving remarkable editing performance in a training-free manner. In terms of the time dimension, they only consider consistency, however, overlook the temporal dynamics of a video involving significant semantic variations. For instance, Figure \ref{fig:framework-related} and its caption describe typical failure cases edited by an existing SOTA method, VidToMe \citep{DBLP:conf/cvpr/LiMYY24}.

\begin{figure*}[t]
        \centering
        \includegraphics[width=1.0\linewidth]{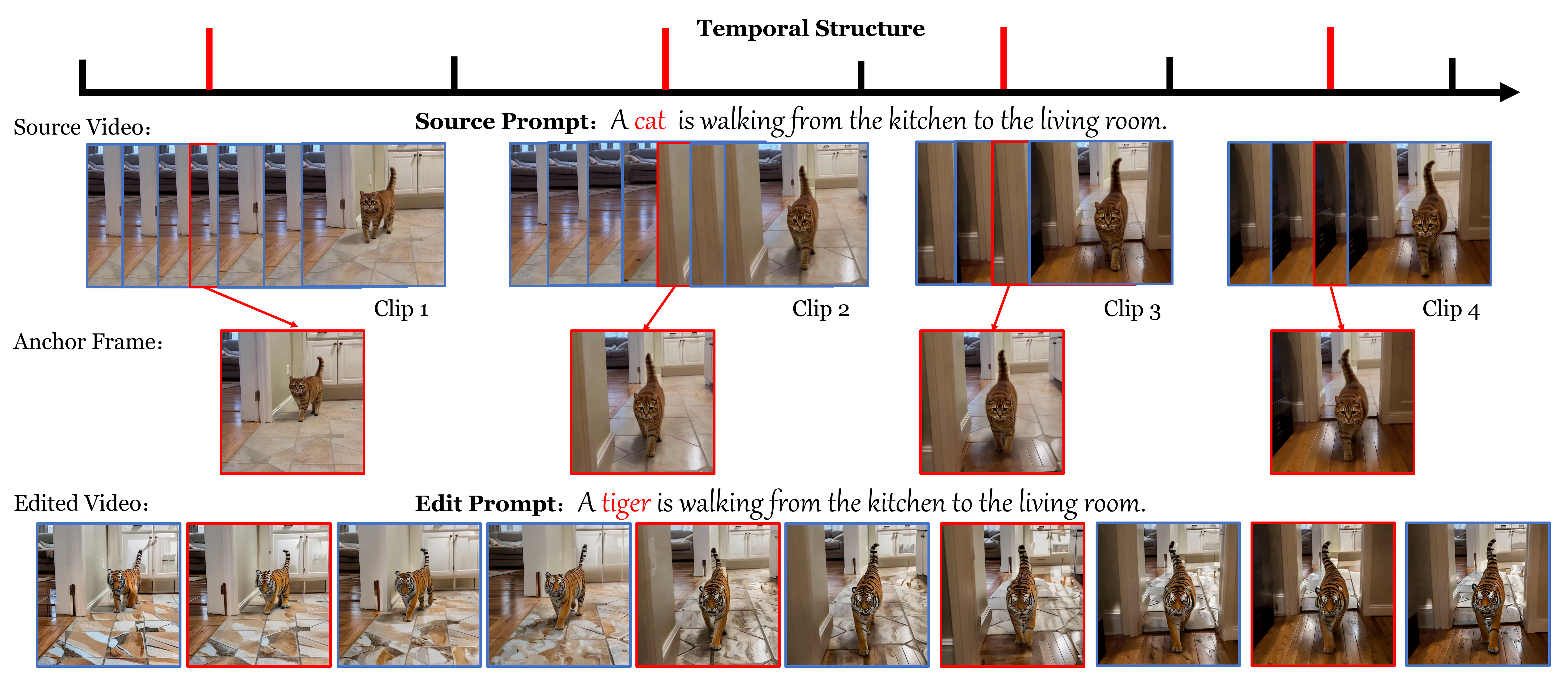}
        \caption{Our method first captures the temporal structure of the source video and then uses it to guide the joint frame editing in the denoising process of the pre-trained diffusion model, achieving preservation of original temporal structure in the edited video.}
        \label{fig:figure1}
\end{figure*}

In this work, we propose a novel zero-shot video editing framework (based on T2I diffusion models) that enables preserving the temporal structure of the source video across generated frames, i.e., Temporal Structure Preservation Video Editing (TSPVE). 
Considering that a long natural video often involves time-dimension semantic variations such as scene or object changes over time, we adaptively partition the source video into variable-length clips of different semantic contents, based on inter-frame similarity measurement by diffusion features (DIFT) \citep{DBLP:conf/nips/TangJWPH23}. 
Meanwhile, for each clip, we anchor a keyframe that best represents it. We refer to this arrangement of clips and anchor frames as the \textit{temporal structure} of a video, as illustrated in Figure \ref{fig:figure1}.

Given the temporal structure of a source video, we introduce the indexes of clips and anchors to organize the joint processing of latent frames clip-by-clip in each denoising diffusion timestep. For each clip, we adopt the token merging technique \citep{DBLP:conf/cvpr/LiMYY24,DBLP:conf/iclr/BolyaFDZFH23} to reduce the memory/computation consumption and enhance the short-term consistency. We view the anchor frame as a destination and merge the temporally redundant transformer tokens to facilitate subsequent joint self-attention. To estimate a merging ratio adaptively for each clip, we constrain token interaction via spatiotemporal windows and use L1 distance to calculate token similarity, filtering highly redundant tokens. Hence, the adaptive ratio mechanism controls the extent of merging for every clip to maintain an equal degree of redundancy.

To enhance the inter-clip continuity, our method considers the information interaction between clips by combining merged tokens of in-pair (adjacent) clips to perform joint self-attention at later diffusion timesteps. It is because the earlier timesteps aim to form different semantic outlines of clips whilst the later ones focus on expressing the details such as style or texture which could be common across clips. For the pairing arrangement of adjacent clips, we design a simple yet effective strategy to make the layout alternate between ($clip_1$,$clip_2$), ($clip_3$,$clip_4$),... and ($clip_1$), ($clip_2$,$clip_3$), ($clip_4$,$clip_5$)... at odd and even time steps.
Our approach can be seamlessly integrated with existing controlling mechanisms (e.g., attention control based \citep{DBLP:conf/iclr/HertzMTAPC23} or ControlNet based \citep{DBLP:conf/iccv/ZhangRA23}), enabling transferring the advantages in image-based editing into video domain. Extensive experiments demonstrate that our training-free method outperforms state-of-the-art approaches in both effectiveness and efficiency. 
The the main contributions of this work are three-fold: 

\begin{itemize}

\item We present a novel zero-shot framework based on pre-trained T2I diffusion models for efficient editing of long videos that involve large temporal semantic variations. We design an adaptive video partition and anchor frame selection scheme to capture the temporal structure describing the variations. And we are the first to consider preserving temporal structure of source video in the video editing task.

\item Given the temporal structure of the source video, we perform clip-adaptive token merging based on spatiotemporal windows to filter redundant tokens to evaluate the token redundancies of counterpart clips within the source video, thereby ensuring similar semantic dynamics in the generated video.

\item We propose an alternated combination strategy to perform joint self-attention of merged tokens of in-pair adjacent clips at later diffusion timesteps, allowing for both semantic discrepancies of edited clips and their continuity. This strategy enhances temporal coherence without requiring large computation across all the clips at one time.

\end{itemize}

\input{related}

\input{method.tex}

\input{experiment}

\section{Conclusion}

This paper proposes a novel zero-shot video editing approach based on text-to-image diffusion models, addressing the issue that existing methods only consider inter-frame temporal consistency while overlooking the temporal structure preservation for the source video. 
Our approach for the first time defines and captures the temporal structure of the original video, and preserves this structure across the generated frames. Under the guidance of the temporal structure, we perform clip-adaptive token merging across frames within each clip for joint self-attention during diffusion process to enhance intra-clip consistency, and adopt an alternated combination strategy for adjacent clips to facilitate inter-clip continuity. Experiments on public video datasets demonstrate that our method outperforms state-of-the-art zero-shot editing approaches in both temporal structure preservation and editing fidelity.

Our method can be seamlessly integrated with diverse image editing control mechanisms without additional training or fine-tuning, while maintaining a trade-off between performance and computational efficiency. However, we acknowledge the limitations of our current work, such as the empirical-based adaptive clip partitioning and lack of multi-level structure preservation. We hence identify several promising directions for future research: 1) further improve the generalization of our method in complex real-world scenarios, exploring multi-level temporal structure modeling for narrative long videos; 2) expand the dedicated quantitative evaluation mechanisms for more complex temporal structure preservation, facilitating in-depth research on structure-aware video generation/editing in the community.

\section*{Authorship contribution statement}
\textbf{Deyin Liu}: Conceptualization, Methodology, Writing draft and Revising. \textbf{Yisheng Ding}: Co-writing draft, Curating data and Validation. \textbf{Zhe Jin}: Investigation and Supervision. \textbf{Xiatian Zhu}: Discussion and Revising. \textbf{Anjan Dutta}:  Guidance and Review. \textbf{Lin Wu}: Review and Editing.

\section*{Declaration of competing interest}
The authors declare that they have no known competing financial interests or personal relationships that could have appeared to
influence the work reported in this paper.

\section*{Acknowledgments}
Thank the support of China Scholarship Council (CSC) for providing research fellowship for the first author. This work was also supported in part by the National Natural Science Foundation of China under Grant 62372150.

\bibliographystyle{elsarticle-num}
\bibliography{elsarticle-num-namesref}
\end{document}

%% file: related.tex
\section{Related Work}
\label{sec:related_work}

\subsection{Diffusion-based Image and Video Editing}
In the context of text-to-image (T2I) diffusion models-based image editing, apart from relying on text prompt to guide the generation of new content, a variety of control mechanisms have been introduced to preserve essential aspects of the original image, such as spatial structure. Notable examples include ControlNet \citep{DBLP:conf/iccv/ZhangRA23}, Plug-and-Play \citep{DBLP:conf/cvpr/TumanyanGBD23}, Self-guidance \citep{DBLP:conf/nips/EpsteinJPEH23}, and pix2pix-zero \citep{DBLP:conf/siggraph/ParmarS0LLZ23},Prompt-to-Prompt\cite{DBLP:conf/iclr/HertzMTAPC23}, and Null-text inversion\cite{DBLP:conf/cvpr/MokadyHAPC23}.
However, when these techniques are extended to video editing, the frame-by-frame processing fails to ensure temporal consistency, leading to visual artifacts or flickering across frames. To address this, recent approaches have adapted T2I diffusion models to the spatiotemporal domain, incorporating architectural modifications to better maintain temporal consistency. Video Diffusion Models (VDMs) \citep{VDM} establish a foundational framework by decomposing the diffusion process into spatial and temporal components. 

Dreamix enhances VDMs through hybrid fine-tuning: it initializes from degraded input videos to retain low-level visual features and fine-tunes on shuffled frames for improved adaptability to text prompts.

Tune-A-Video \cite{DBLP:conf/iccv/WuGWLGSHSQS23} proposes efficient video adaptation by incorporating sparse causal attention and freezing most model parameters. 
Video-P2P \cite{DBLP:conf/cvpr/LiuZ00J24} extends image editing techniques to video domain by employing frame-wise interaction for edit propagation. 
However, these methods often need retraining or fine-tuning, or optimization, which involve a lot of costly computation. Recently, a category of zero-shot video editing methods have been very popular among researchers due to its training-free manner while presenting remarkable performance.

\subsection{Zero-shot Video Editing}
Zero-shot video editing frameworks have revolutionized video manipulation by leveraging pre-trained T2I diffusion models for flexible and efficient text-guided modifications without fine-tuning. 
Pix2Video \citep{DBLP:conf/iccv/CeylanHM23} integrates sparse-causal attention and latent guidance, using original image predictions as denoising proxies; FateZero \citep{DBLP:conf/iccv/QiCZLWSC23} emphasizes spatio-temporal preservation via attention features during inversion for superior motion retention; 
Text2Video-Zero \citep{DBLP:conf/iccv/Text2Video-Zero} employs cross-frame attention, initial frame integration, and background smoothing for video synthesis; 
Rerender-A-Video \citep{DBLP:conf/siggrapha/RerenderAVideo,DBLP:conf/cvpr/YangZLL24} imposes hierarchical cross-frame constraints.
Besides, VidToMe \citep{DBLP:conf/cvpr/LiMYY24} and TokenFlow \citep{DBLP:conf/iclr/GeyerBBD24} stand out for enhancing temporal consistency through transformer token compression and propagation respectively. However, their uniform video partition and random keyframe sampling operations focus only on global consistency in temporal dimension, ignoring the preservation of the temporal structure of the source video\cite{DBLP:conf/iclr/CongXSCRXPR0024,DBLP:conf/icml/CohenKKHM24}. Our proposed method addresses this by adaptive video partition and anchor frame selection according to the temporal semantic variations, performing clip-adaptive token merging with the temporal structure of source video.

%% file: method.tex
\section{Preliminaries}\label{sec:preliminary}
\subsection{Latent Diffusion Models}
Latent diffusion models (LDM) \citep{DBLP:conf/cvpr/BlattmannRLD0FK23} perform diffusion processes in a latent space with dimension reduced by an Autoencoder such as VAE \Citep{DBLP:journals/corr/KingmaW13}. 
In the latent space of LDM, we generally employ the DDIM inversion \citep{DBLP:conf/iclr/SongME21} to invert the clean latents of source video frames to their corresponding noisy latents. That is, by reversing the deterministic DDIM sampling process, we derive the latent trajectory \( \{z_t^i\}_{t=0}^T \) for each frame \( i \), making that the acquired noisy latent \( z_T^i \) a good noise seed for subsequent generative editing using DDIM sampling.

The forward process corrupts clean image latents \(z_0\) into noisy latents \(z_t\) , while the reverse process denoises \(z_T\) to \(z_0\) guided by text prompts.
Forward Process: 
A clean latent \( z_0 \) is iteratively perturbed by Gaussian noise as follows:
\begin{equation}
    z_t = \sqrt{\alpha_t} z_0 + \sqrt{1 - \alpha_t} \epsilon, \quad \epsilon \sim \mathcal{N}(0, \mathbf{I})
\end{equation}
where \(\alpha_t\) is a noise schedule parameter that decreases with the timestep \( t \). This process gradually transforms the latent into a random noise distribution \( z_T \).
Reverse Denoising Process: 
Starting from \( z_T \), the diffusion model \(\epsilon_\theta\) predicts the noise at each timestep \( t \), updating \( z_t \) to \( z_{t-1} \):
\begin{equation}
    z_{t-1} = \sqrt{\frac{\alpha_{t-1}}{\alpha_t}} z_t + \left( \sqrt{\frac{1 - \alpha_{t-1}}{\alpha_t}} - \sqrt{\frac{1 - \alpha_t}{\alpha_t}} \right) \epsilon_\theta(z_t, t, \mathcal{C}),
\end{equation}
where \(\mathcal{C}\) denotes the text prompt embedding. This iterative process leverages pre-trained T2I models like Stable Diffusion \citep{DBLP:conf/cvpr/BlattmannRLD0FK23}, Glide\citep{DBLP:conf/icml/NicholDRSMMSC22} and rectified flow transformers to progressively generate high-quality frames aligned with the editing prompt.
This allows the diffusion model to edit the video while preserving temporal consistency.

\subsection{Video Token Merging and Unmerging}
Token Merging (ToMe) \citep{DBLP:conf/iclr/BolyaFDZFH23} is a technique designed to increase the throughput of Vision Transformer (ViT) models by progressively merging redundant tokens within transformer blocks. It combines similar tokens to reduce the redundancy as well as the number of tokens, speeding up the computation. 
Given a set of input tokens \( T_{\text{in}} \in \mathbb{R}^{B \times N \times C} \) (where \( B \) is the number of frames, \( N \) is the token count per frame, and \( C \) is the feature dimension), the algorithm first divides $T_{\text{in}}$ into a destination (\textit{dst}) set \( T_{\text{in}}^{\text{dst}} \) (typically, consisting of the $N$ tokens of a sampled keyframe) and a source (\textit{src}) set \( T_{\text{in}}^{\text{src}} \) (including \((B - 1) \cdot N\) tokens of the rest frames), and computes pairwise cosine similarities between \textit{src} and \textit{dst} tokens.  
Then the top \( r \) most similar token pairs are linked via edges to form a matching map \( \text{Match}(\textit{src},\textit{dst},r) \), which guides the merging operation:
\begin{equation}
    T_{\text{Merged}} = \text{Merge}(T_{\text{in}}, \text{Match}(\textit{src},\textit{dst},r)) \label{eq1}
\end{equation} 
where \( \text{Merge}(\cdot) \) means concatenating all the tokens from both sets and deleting the tokens from the \textit{src} side of the \( r \) most similar token pairs rather than averaging the values of each pair.
The merged tokens will go through the self-attention module of transformer block, and then perform unmerging according to the matching map again. The missing tokens in the \textit{src} set will be replaced with their edge-linked ones in the \textit{dst} set:    
\begin{equation}
    T_{out} = \text{Unmerge}(T^{'}_{\text{Merged}}, \text{Match}(\textit{src},\textit{dst},r)) \label{eq2}
\end{equation}
where $T^{'}_{\text{Merged}}$ denotes the updated tokens via self-attention module.

\section{Proposed Method} \label{sec:Proposed Method}

\begin{figure*}[t]
        \centering
        \includegraphics[width=1.0\linewidth]{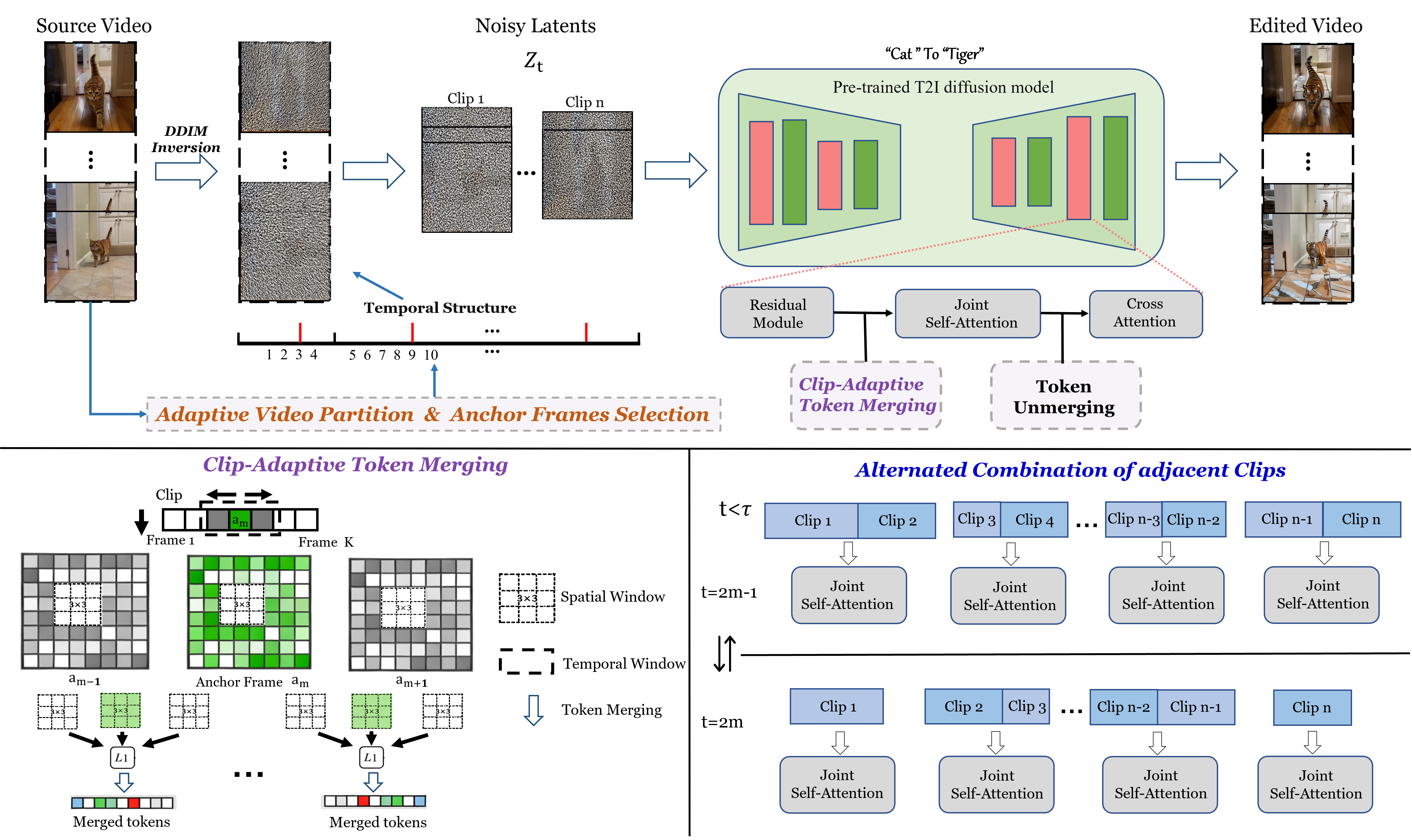}
        \caption{\textbf{Overview of the proposed video editing method.} We first capture the source video's temporal structure via Adaptive Video Partition and Anchor Frame Selection. Guided by the structure, a pre-trained T2I diffusion model progressively denoises DDIM-inverted noisy latents to generate new frames. In each timestep, clip-adaptive token merging/unmerging around self-attention enhances short-term consistency and reduce memory/computation cost. For later timesteps (\(t<\tau\)), adjacent clips are alternately combined for joint self-attention at odd/even timesteps, boosting inter-clip continuity.}
        \label{fig:framework}
\end{figure*}

We aim to perform training-free video modification guided by a textual edit prompt, while preserving key source characteristics such as spatial structure and temporal dynamics. 
In spatial dimension, we integrate existing controlling mechanisms of image editing into the diffusion generation process for editing the frames to preserve the spatial structure of source frames. 
In the temporal dimension, we propose to preserve the temporal structure of the source video across the generated frames.
To this end, we need to capture the temporal structure of the original video and embed it into the denoising diffusion process of the generative-based video editing. The overview of the proposed method is shown in Figure \ref{fig:framework}. 

\subsection{Zero-Shot Video Editing Pipeline}
Given a source video $V$ of $n$ frames \([I^1, \dots, I^n]\), we employ an autoencoder VAE \Citep{DBLP:journals/corr/KingmaW13} to encode the frames into the latent sequence $[z^1_0, \dots, z^n_0]$ . Then DDIM inversion \citep{DBLP:conf/iclr/SongME21} is applied to turn the clean latent frames to noisy ones $[z^1_T, \dots, z^n_T]$ via $T$ time steps. 
Starting from the inverted noisy latent frames, we iteratively denoise them by a pre-trained T2I diffusion model $\epsilon_{\theta}$ conditioned on the textual edit prompt $\mathcal{P}$, to progressively realize targeted video editing. 
For a long video with large temporal semantic variations, we adaptively partition the source video into variable-length clips of different semantic contents and identify an anchor frame for each clip, and thus acquire the temporal structure. 
Then under the guidance of the temporal structure, before the self-attention module of each layer, we perform clip-adaptive token merging via spatiotemporal windows and L1-distance-based similarity filtering in each diffusion time step. 
After self-attention, we restore the updated tokens to the original size by token unmerging. Through $T$ iterations of denoising, we obtain the latent embeddings $[{z'}^1_0, \dots, {z'}^n_0]$ which are then decoded by VAE\Citep{DBLP:journals/corr/KingmaW13} into the edited video $V'$. Besides, We propose an alternated combination strategy to perform joint self-attention of merged tokens across adjacent clips from some late diffusion timestep on, which ensures the inter-clip continuity and smooth appearance in the new video.  

\begin{algorithm}[t]
\footnotesize  
\caption{Temporal structure capture}\label{alg:temporal_structure_capture}
\begin{algorithmic}[1]
\setlength{\topsep}{0pt}   
\setlength{\partopsep}{0pt}  
\REQUIRE 
    $[\mathbf{F}^1,\dots,\mathbf{F}^n]$: DIFT features, $H,W,l,s,m_s,w_s$: parameters
\ENSURE
    $\mathcal{S}=\{s_m\}$: Clip start indices, $\mathcal{A}=\{a_m\}$: Anchor frames

\STATE $\mathcal{S}\leftarrow\{1\}, i\leftarrow1, j\leftarrow2$
\WHILE{$j \leq n$}
    \STATE Compute similarity heatmap $\mathbf{H}_{i,j}$
    \IF{$\text{mean}(\mathbf{H}_{i,j})<m_s$ or $\exists p:\mathbf{H}_{i,j}(l,s,p)<w_s$}
        \STATE $\mathcal{S}\leftarrow\mathcal{S}\cup\{i\}, i\leftarrow j, j\leftarrow i+1$
    \ELSE
        \STATE $j\leftarrow j+1$
    \ENDIF
\ENDWHILE

\FOR{each clip $m$}
    \STATE $K_m\leftarrow s_{m+1}-s_m$
    \FOR{$k=1$ to $K_m$}
        \STATE $\mathbf{f}_m^k\leftarrow\text{Average}(\mathbf{F}^{s_m+k-1}_m)$
    \ENDFOR
    \STATE $\mathcal{C}_m\leftarrow$ cosine similarity matrix of $\{\mathbf{f}_m^k\}_{k=1}^{K_m}$
    \STATE $a_m\leftarrow s_m+\arg\max_i\sum_{j=1}^{K_m}\mathcal{C}_m^{i,j}-1$
    \STATE $\mathcal{A}\leftarrow\mathcal{A}\cup\{a_m\}$
\ENDFOR
\RETURN $\mathcal{S}, \mathcal{A}$
\end{algorithmic}
\end{algorithm}

\subsection{Adaptive Video Partition and Anchor Frame Selection}\label{section:4_1}

To capture the temporal structure of the original video, we first conduct Adaptive Video Partition (AVP) to split  the source video into variable-length clips with distinct semantic contents according to temporal semantic dynamics, and then perform adaptive Anchor Frame Selection (AFS) by exploring inter-frame similarities within each clip. The detailed procedures of AVP-AFS are presented in Algorithm 1.

To be specific, based on DIFT \citep{DBLP:conf/nips/TangJWPH23}, we leverage a pre-trained diffusion network \(\epsilon_{\theta}\) to extract diffusion features for a source video,denoted as $ [\mathbf{F}^1, \dots, \mathbf{F}^n]$, where \(\mathbf{F}^i \in {R}^{C \times H \times W}\) represents the feature of the $i$-th frame, $C$ is the feature dimension, and \(H \times W\) denotes the spatial resolution of every frame. 
Following \citep{DBLP:journals/corr/abs-2502-05433}, we compute the token-wise cosine similarity of DIFT features to acquire a temporal dynamics heatmap $\mathbf{H}_{i,j} \in R^{H \times W}$  between the $i$-th and the $j$-th frames. We traverse the frame sequence and calculate the similarity heatmap between the current frame $l$ and the start frame $S_m$ of the current clip $m$. If the mean value of the heatmap between a pair of frames is lower than a pre-defined threshold $h_m$, i.e., $\text{Mean}(H_{S_m,l})<h_m$,

the current frame is marked as the start of a new clip. We repeat the operation sequentially from the new starting point until the end of the video. In other words, we define the semantics inflection points as the clip boundaries, where the index set $\mathcal{S} = \{s_1, ..., s_M\}$ consists of the starting points of $M$ clips. 

To identify an anchor frame for each clip, we first derive a compact representation for the $i$-th frame by performing \textit{average} operation on the spatial dimensions (\(H \times W\)) to compress \(\mathbf{F}^i\) into a single $C$-dimensional vector $\mathbf{f}^i$.

For clip $m$ containing \(K_m\) frames, we concatenate the frame-wise representations to form a matrix \(\mathcal{F}_m = [ \mathbf{f}_m^1,\mathbf{f}_m^2,...,\mathbf{f}_m^{K_m} ] \in {R}^{C \times K_m}\), where each column corresponds to a frame’s \(C\)-dimensional representation. By using \(\mathcal{F}_m\), we construct a cosine similarity matrix \(\mathcal{C}_{m} \in {R}^{K_m \times K_m}\) to measure pairwise between-frame similarity:  
\begin{equation}
   \mathcal{C}^{i,j}_m = \frac{\mathbf{f}_m^i \cdot \mathbf{f}_m^j}{\|\mathbf{f}_m^i\| \cdot \|\mathbf{f}_m^j\|}.
\end{equation}
The correlation score of the \(i\)-th frame, measuring its correlations to all other frames in the clip, is defined as the row sum of the $i$-th row of \(\mathcal{C}_m\):  
$c_m^i = \sum_{j=1}^{K_m} \mathcal{C}_m^{i,j}$. Then the frame with the highest correlation score is selected as the anchor frame \(a_m\) for clip \(m\). These adaptively selected anchor frames best represent the clips. With the indexes of the video clips and their anchors, we capture the temporal structure of the source video.

\subsection{Clip-Adaptive Token Merging} \label{section:4_2}

To reconcile latent token merging efficiency with source video temporal structure preservation in the denoising process of zero-shot video editing, and mitigate the inflexibility of traditional token merging with fixed parameters for video semantic dynamics, we propose a clip-adaptive token merging strategy based on AVP-AFS outputs. This strategy constrains token interaction within local spatiotemporal windows, filters redundancy via similarity measurement, and preserves the semantic dominance of anchor frames, enabling efficient intra-clip
token merging while retaining core semantic information.

Specifically, the destination token set (\(T_{\text{dst}}\)) is defined as all spatial tokens of the AVP-AFS-selected anchor frame, where the token \(Z_{a_m}(x,y)\) at spatial coordinate \((x,y)\) serves as the merging benchmark  for regional tokens in the clip to center merging on core semantics. Leveraging video local semantic continuity, we design a constrained spatiotemporal window for token interaction: a 3×3 spatial neighborhood window to avoid cross-object/scene invalid merging, and a temporal window with fixed radius k=1 (covering \([a_m-1, a_m, a_m+1]\)) for frame-level interaction.
Mirror padding is applied to clip edge frames via coordinate mapping to eliminate zero-padding noise and maintain window structural consistency, with valid merge candidates for each token \(Z_{f,x,y}\) restricted to the spatiotemporal window intersection.

An anchor frame-centered iterative merging workflow is designed for semantic transfer: we first merge tokens in the core 3-frame temporal window, then take the merged result as a new benchmark to iteratively expand merging to all outer frames of the clip, with each frame participating in merging exactly once to realize efficient semantic transfer from the anchor frame to the entire clip without redundant computation. Aligned with the training-free nature of zero-shot editing, we adopt the L1 distance as the similarity metric (calculated only within spatiotemporal windows) to filter highly redundant tokens, avoiding the complex norm and square root calculations of cosine similarity. The normalized similarity coefficient is defined as: 
\begin{equation}
sim(Z_{f,x,y}, Z_{f',x',y'}) = 1 - \frac{1}{C} \sum_{c=1}^C \left| Z_{f,x,y,c} - Z_{f',x',y',c} \right|
\end{equation}
where, \(Z_{f,x,y,c}\) denotes the value of token \(Z_{f,x,y}\) in the c-th feature dimension, and $C$ is the feature dimension. Only tokens with \(sim > \tau_{\text{sim}}\) are retained as merging candidates.

To balance anchor frame semantic dominance and merged token feature integrity, a soft-weighted merging strategy is proposed, assigning a fixed weight of 1 to anchor frame tokens to preserve core semantics and weighting candidate tokens by their similarity coefficients. The merging formula is:
\begin{equation}
Z_{\text{merge}}(x,y) = \frac{Z_{a_m}(x,y) + \sum_{(f',x',y') \in W} sim \cdot Z_{f',x',y'}}{1 + \sum_{(f',x',y') \in W} sim}
\end{equation}
where W represents the set of valid spatiotemporal window candidates satisfying  \(sim > \tau_{\text{sim}}\).. A mapping index table \(M \in \mathbb{R}^{B \times N_s \times N_s \times 2}\) is synchronously generated during merging:  \(M[f,x,y,0]\)  records the spatial coordinates of matched anchor frame tokens for accurate unmerging, and  \(M[f,x,y,1]\) stores merging weight coefficients for reverse feature distribution. This table is also reused by the ACAC module to quickly locate cross-clip token correspondences, reducing feature matching overhead for inter-clip joint self-attention.

Notably, this clip-adaptive design achieves a significant reduction in computational complexity compared with the chunk merging in VidToMe\citep{DBLP:conf/cvpr/LiMYY24}, as it avoids global token similarity calculations and restricts interaction to local spatiotemporal windows, realizing efficient token merging while adapting to the semantic dynamics of different clips.

\subsection{Alternated Combination of Adjacent Clips} \label{subsec_4_3}

For every clip, the merged tokens go through the self-attention module and then the \textit{src} and \textit{dst} sets are restored to their respective original size via token unmerging. Such intra-clip feature interaction and update through multiple layers of transformer blocks of multiple timesteps enhance the short-term consistency of generated frames. However, they overlooks the long-term consistency, leading to inter-clip discontinuity of edited video. Recall that the earlier diffusion time steps focus on the shaping of semantic outlines while the later time steps stress detail rendering, to allow for both the clips' respective semantics and appearance' smoothness in the generated video, as shown in Figure \ref{fig:framework}, we perform joint self-attention of merged tokens of every two adjacent clips only at late-stage diffusion timesteps (When $t<\tau$) and alternate the neighbor-combination layouts of clips between odd and even timesteps. This alternation strategy enhances long-term consistency through a sequence of in-pair neighboring interactions  reducing significant computation and memory footprint compared to the global token merging in VidToMe \citep{DBLP:conf/cvpr/LiMYY24}.

%% file: experiment.tex
\section{Experiments}
\subsection{Experimental Setting}
Our method is based on a pre-trained T2I model integrated with an existing control technique from image editing. Specifically, we select Stable Diffusion (SD) version 1.5 \citep{DBLP:conf/cvpr/RombachBLEO22} as the core image generation component and use DDIM scheduler for sampling and inversion, with the sampling steps set as T=50. To show our method's ability to edit long videos with small memory footprint and computation cost, we run it on an NVIDIA RTX 4090 GPU with mixed-precision inference. 
Our approach integrates two different image control techniques including Plug and Play (PnP) \citep{DBLP:conf/cvpr/TumanyanGBD23} and ControlNet architecture \citep{DBLP:conf/iccv/ZhangRA23} to control the generation of frames to preserve some source information such as spatial structure. For Anchor Frame Selection, we empirically determine the threshold $ h_m $ as 0.6 by conducting a grid search on the LongV-EVAL validation set.
For Clip-Adaptive Token Merging, in the spatial dimension, a fixed 3×3 neighborhood window is adopted; in the temporal dimension, a fixed window radius of 1 is set, allowing only temporally contiguous frames to participate in token merging. For frames at the clip edges that cannot form a complete spatiotemporal window, mirror padding is implemented through coordinate mapping to avoid noise interference caused by zero-padding. A similarity threshold of \(\tau_{\text{sim}}\)= 0.8 is defined, and only highly redundant tokens with a normalized similarity coefficient (sim) > 0.8 are retained as merging candidates. During the merging process, the weight of anchor frame tokens is fixed at 1 to ensure the core semantics of the clip dominate, while candidate tokens use their own normalized similarity coefficient (sim) as their weight.
For extracting DIFT from each frame of the source video, we select the features corresponding to t=0 , which are extracted from the intermediate layer of the 2D U-Net Decoder. 
The heatmap threshold $h_m$ for the adaptive video partition is set as 0.6.
We perform the alternated combination strategy for the last 10 diffusion timesteps only ($\tau$=10).  
Given that Tune-A-Video\citep{DBLP:conf/iccv/WuGWLGSHSQS23}, Video-P2P\citep{DBLP:conf/cvpr/LiuZ00J24}, FateZero\citep{DBLP:conf/iccv/QiCZLWSC23}, vid2vid-zero\citep{DBLP:journals/corr/abs-2303-17599}, and TokenFlow\citep{DBLP:conf/iclr/GeyerBBD24} cannot edit long videos in a single inference, we segment such videos for editing. Based on their computational resource requirements, we process 64, 32, and 16 frames per batch respectively.

\paragraph{Datasets} 
Following \citep{DBLP:conf/iccv/QiCZLWSC23,DBLP:conf/cvpr/LiMYY24}, we use videos from the LongV-EVAL\citep{DBLP:journals/corr/abs-2502-05433} dataset, MiraData\citep{DBLP:conf/nips/Ju00YWZX0S24} dataset and other in-the-wild videos as source videos, covering various subjects, including people, animals and landscapes. The prompts for these videos are obtained using GPT-3.5 \citep{ray2023chatgpt}, adapted from prior studies, or self-created by the authors.The LongV-EVAL\citep{DBLP:journals/corr/abs-2502-05433} dataset contains 75 videos, each approximately 1 minute long, with a frame rate of 20-30 fps. The content covers multiple themes, including people, animals, and landscapes. The MiraData\citep{DBLP:conf/nips/Ju00YWZX0S24} features long videos with an average duration of 72 seconds and structured subtitles of 6 types, averaging 318 words.During the editing process of the LongV-EVAL\citep{DBLP:journals/corr/abs-2502-05433} dataset,MiraData\citep{DBLP:conf/nips/Ju00YWZX0S24} dataset,the video resolution was set to 384×672.
To comprehensively evaluate our method's performance, we design three types of video editing experiments: object replacement, background replacement, and style transformation.

\paragraph{Metrics} 
To systematically quantify the performance of our method, we adopt a comprehensive set of evaluation metrics informed by prior works \citep{DBLP:conf/iccv/CeylanHM23,DBLP:conf/iccv/QiCZLWSC23}, encompassing both objective measurements and subjective judgment.Specifically, we leverage established numerical indicators that assess the smoothness and fidelity of edited video sequences over time. The Peak Signal-to-Noise Ratio (PSNR) \citep{DBLP:conf/cvpr/JiangSJ0LK18} quantifies pixel-wise fidelity by comparing edited frames against ground-truth sequences. Interpolation Error is employed to measure discrepancies in motion interpolation between consecutive frames. Additionally, Warp Error measures the optical flow-based frame warping, ensuring consistent motion dynamics across the video timeline. The Structural Similarity Index (SSIM) \citep{DBLP:journals/tip/WangBSS04} evaluates the similarity in structure, luminance, and contrast between edited frames and ground truth, offering a perceptually meaningful metric for pixel-level quality. Meanwhile, the Learned Perceptual Image Patch Similarity (LPIPS) \citep{DBLP:conf/cvpr/ZhangIESW18} gauges the perceptual difference between edited and ground-truth frames using deep convolutional features, which correlates closely with human visual judgment.To assess the degree of alignment between visual output and input text prompt, we utilize two CLIP-based metrics. The Frame CLIP Score computes cosine similarity between visual features of individual edited frames and text-encoded representations using the CLIP model, providing a fine-grained measure of visual-text consistency at the frame level. Concurrently, the Text CLIP Score assesses global coherence by comparing aggregated video features with the input text prompt.Furthermore, a user study is conducted. Ten users compare and select their preferred videos among videos edited by the baselines and our method, with the vote rate serving as the final metric for user preference.

To directly quantify the preservation of the source video’s high-level temporal structure, we propose a dedicated Temporal Structure Preservation Rate (TSPR) metric built on the official Subject Consistency (SC) and Background Consistency (BC) metrics from the widely adopted VBench benchmark\citep{DBLP:conf/cvpr/HuangHYZS0Z0JCW24}. SC and BC characterize the long-range temporal consistency of the primary subject and background across the video sequence, respectively, which are the core components of a video’s temporal structure. The formal definition of TSPR is:

\begin{equation}
\text{TSPR}=\frac{1}{2} \times\left( \frac{SC(C_{edit})}{SC(C_{ori})} + \frac{BC(C_{edit})}{BC(C_{ori})} \right)
\end{equation}

where \(C_{ori}\) and \(C_{edit}\) denote the full original source video and edited video, respectively. TSPR ranges from 0 to 1, with a value closer to 1 indicating better retention of the source video’s temporal structure after editing. To ensure statistical robustness, we set a valid lower bound of 0.2 for \(SC(C_{ori})\) and \(BC(C_{ori})\): terms below this threshold are excluded from the calculation to avoid metric distortion, while remaining fully compatible with the VBench\citep{DBLP:conf/cvpr/HuangHYZS0Z0JCW24} metric system.

\paragraph{Baselines} 
This study performs experimental comparisons with two primary categories of video editing methods: one-shot approaches based on fine-tuning, including Tune-A-Video \citep{DBLP:conf/iccv/WuGWLGSHSQS23} and Video-P2P \citep{DBLP:conf/cvpr/LiuZ00J24}, as well as zero-shot approaches such as FateZero \citep{DBLP:conf/iccv/QiCZLWSC23}, vid2vid-zero \citep{DBLP:journals/corr/abs-2303-17599}, TokenFlow \citep{DBLP:conf/iclr/GeyerBBD24}, VidToMe \citep{DBLP:conf/cvpr/LiMYY24}, AdaFlow \citep{DBLP:journals/corr/abs-2502-05433}, ControlVideo(depth) \citep{DBLP:conf/iclr/Zhang0J0Z024}, RAVE \citep{DBLP:conf/cvpr/KaraKYRY24}, Slicedit \citep{DBLP:conf/icml/CohenKKHM24}, and VideoGrain \citep{DBLP:conf/iclr/YangZF025}. Notably, RAVE is initially designed for image editing tasks, and we adopt their official open-source implementations and adapt them to the video editing scenario for fair comparison. Among these, VidToMe \citep{DBLP:conf/cvpr/LiMYY24} and AdaFlow\citep{DBLP:journals/corr/abs-2502-05433} serve as the key baselines due to their significant contributions to the domain of zero-shot video editing, and ControlVideo, RAVE, Slicedit, VideoGrain are also included as representative state-of-the-art baselines to fully validate the advancement of our method, which directly align with the focus of our study.
Similar to VidToMe\citep{DBLP:conf/cvpr/LiMYY24}, we integrate two prominent image control techniques—Plug and Play (PnP) \citep{DBLP:conf/cvpr/TumanyanGBD23} and ControlNet architecture \citep{DBLP:conf/iccv/ZhangRA23}—into our experiments to enable more nuanced video editing through the incorporation of external control mechanisms. These control techniques are integrated to enhance the flexibility and effectiveness of video editing, and their interaction with our method provides valuable insights. Other methods, however, do not easily support the direct and flexible integration of image editing control mechanisms into video editing tasks, which sets our work apart.
For all baseline methods, we use the default settings provided in their respective official implementations, ensuring consistency and fairness in the comparison. Since VidToMe\citep{DBLP:conf/cvpr/LiMYY24}, TokenFlow\citep{DBLP:conf/iclr/GeyerBBD24} and so on are unable to handle long video editing in a single inference due to computational limitations, we segment the long videos into smaller clips for processing. Based on the computational resource requirements of these methods, we edit 128, 32, and 16 frames at a time, respectively. This segmentation strategy ensures that each method can operate within feasible resource constraints while maintaining the quality of the edits.

\subsection{Main Results}

\begin{figure*}[t]
    \centering
        \includegraphics[width=1.0\linewidth]{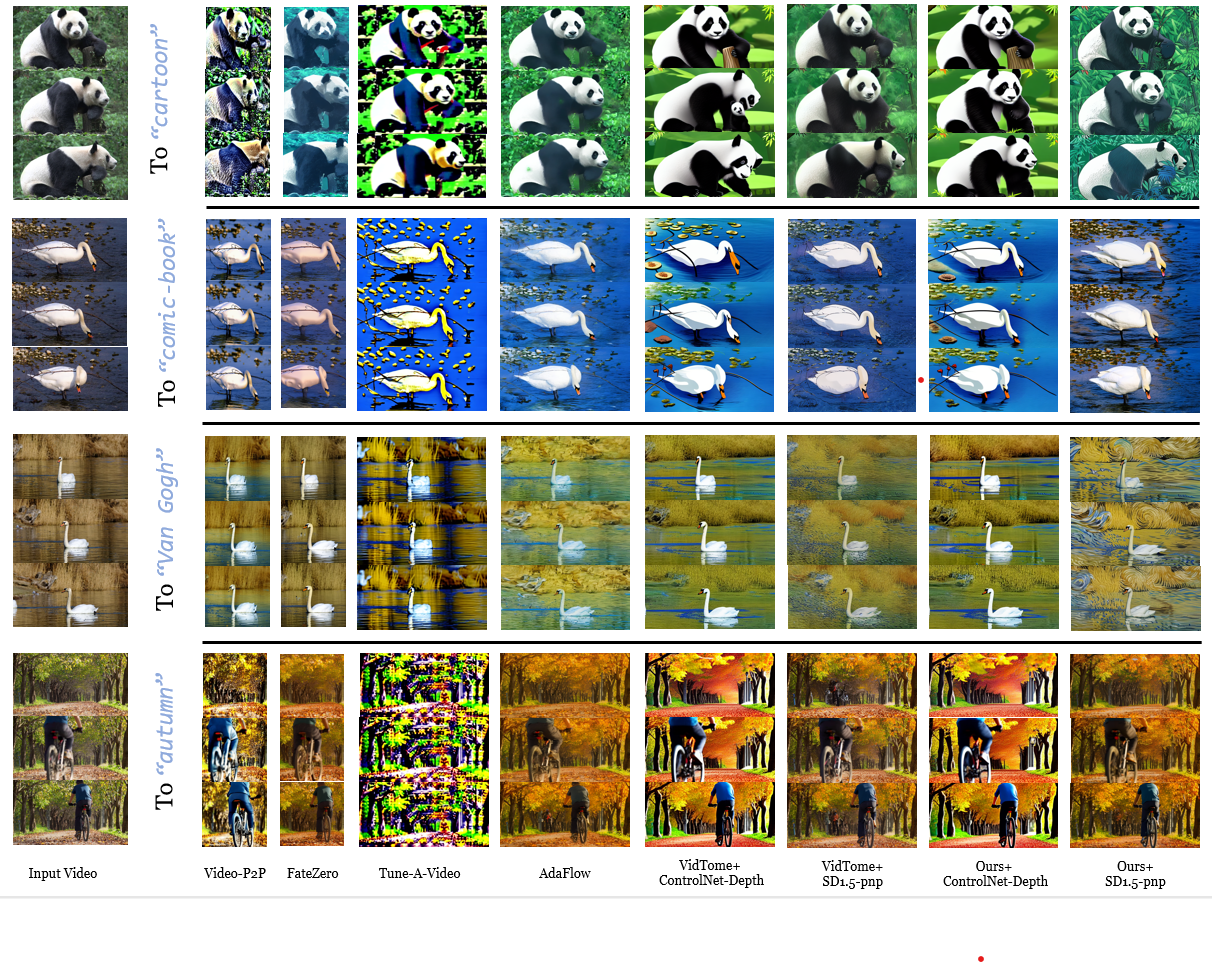}
        \caption{A qualitative comparison of our method with baseline methods on long video datasets. The editing results of our method are consistent over time in global style and local texture, and preserve the temporal semantic dynamics (structure) well.}
        \label{fig:comparison_results_long}  
\end{figure*}

\begin{figure*}[t]
    \centering
        \includegraphics[width=1.0\linewidth]{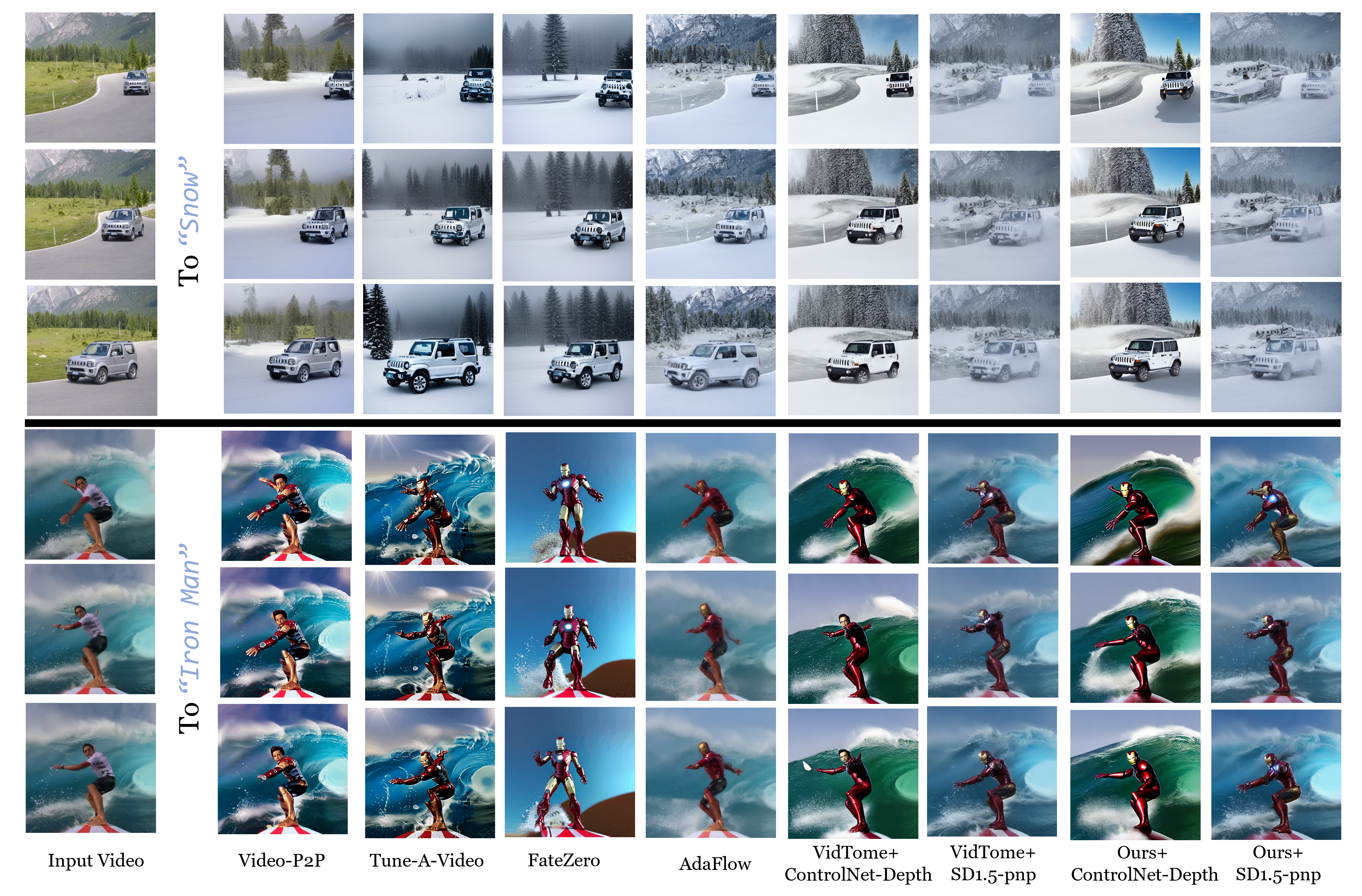}
        \caption{A qualitative comparison of our method with baseline methods on short video datasets.}
        \label{fig:comparison_results_short}  
\end{figure*}
\subsubsection{Qualitative Analysis}
Figures \ref{fig:comparison_results_long} and \ref{fig:comparison_results_short} show sampled frames at identical indices from videos edited by different methods, highlighting discrepancies in spatial/temporal preservation.
For the short-video dataset (Figure \ref{fig:comparison_results_short}), Tune-A-Video\citep{DBLP:conf/iccv/WuGWLGSHSQS23}, Video-P2P\citep{DBLP:conf/cvpr/LiuZ00J24}, and FateZero\citep{DBLP:conf/iccv/QiCZLWSC23} (second to fourth columns) fail to preserve spatial structure, e.g., lost curved grassland-road boundaries, distorted background wave shape/size, and altered human poses/orientations. Temporally, Tune-A-Video\citep{DBLP:conf/iccv/WuGWLGSHSQS23} struggles with simple transitions, Video-P2P\citep{DBLP:conf/cvpr/LiuZ00J24} introduces abrupt background changes/rigid object deformations, and FateZero\citep{DBLP:conf/iccv/QiCZLWSC23} overlooks local details, causing consecutive frame discontinuities. Our method achieve better frame consistency.
For the long-video dataset (Figure \ref{fig:comparison_results_long}),
AdaFlow\citep{DBLP:journals/corr/abs-2502-05433} achieves relatively efficient inference via attention optimization, yet it fails to explicitly model the source video’s intrinsic temporal structure. This results in inconsistent semantic expression across extended frame sequences, with occasional disruptions in motion coherence when handling large temporal variations, such as sudden object appearances or disappearances between consecutive clips. VidToMe, while enhancing intra-frame consistency through global token merging, relies on uniform video partitioning and random keyframe sampling. This rigid partitioning strategy cannot adapt to the non-uniform temporal semantic dynamics of long videos, resulting in compromised temporal structure retention—manifested as blurred semantic boundaries between adjacent clips and reduced consistency in object attributes across the entire sequence. As shown in the last column of Figure \ref{fig:comparison_results_long}, our method maintains consistent global style and fine-grained local texture throughout the long sequence, effectively preserving both the spatial structure and temporal semantic dynamics of the source video.

\subsubsection{Quantitative Analysis}
\begin{table*}[t]
    \centering
    \resizebox{\textwidth}{!}{ 
    \begin{tabular}{lcccccccccccc}  
        \toprule
        & \multicolumn{5}{c}{Temporal Consistency} & \multicolumn{2}{c}{Text Alignment} & \multicolumn{3}{c}{Temporal Structure} & \multicolumn{1}{c}{User Study} & \multicolumn{1}{c}{Running Time} \\
        \cmidrule(lr){2-6} \cmidrule(lr){7-8} \cmidrule(lr){9-11} \cmidrule(lr){12-12} \cmidrule(lr){13-13}  
        & PSNR$\uparrow$ & Inp. Err.$\downarrow$ & Warp Error$\downarrow$ & SSIM $\uparrow$ & LPIPS $\downarrow$ & Frame CLIP$\uparrow$ & Text CLIP$\uparrow$ & SC(edit)$\uparrow$ & BC(edit)$\uparrow$ & TSPR$\uparrow$ & Voting rate$\uparrow$ & per video (min) $\downarrow$ \\  
        \midrule
        Tune-A-Video \citep{DBLP:conf/iccv/WuGWLGSHSQS23} & 24.014 & 0.134 & 0.225 & 0.416 & 0.591 & 0.949 & 0.285 & 0.827 & 0.905 & 0.830 & 0.02 & 180:30 \\
        Video-P2P \citep{DBLP:conf/cvpr/LiuZ00J24} & 25.781 & 0.106 & 0.162 & 0.218 & 0.541 & 0.941 & 0.275 & 0.842 & 0.918 & 0.885 & 0.02 & 128:46 \\
        FateZero \citep{DBLP:conf/iccv/QiCZLWSC23} & 25.225 & 0.164 & 0.198 & 0.404 & 0.533 & 0.948 & 0.301 & 0.842 & 0.918 & 0.855 & 0.02 & 92:34 \\
        vid2vid-zero \citep{DBLP:journals/corr/abs-2303-17599} & 24.514 & 0.118 & 0.179 & 0.422 & 0.553 & 0.957 & 0.270 & 0.860 & 0.935 & 0.870 & 0.02 & 82:35 \\
        TokenFlow \citep{DBLP:conf/iclr/GeyerBBD24} & 25.740 & 0.107 & 0.145 & 0.731 & 0.382 & 0.958 & 0.283 & 0.908 & 0.953 & 0.935 & 0.05 & 14:34 \\
        AdaFlow \citep{DBLP:journals/corr/abs-2502-05433} & 28.277 & 0.063 & 0.145 & 0.750 & 0.365 & 0.959 & 0.280 & 0.927 & 0.968 & 0.957 & 0.12 & 10:55 \\
        ControlVideo(depth) \cite{DBLP:conf/iclr/Zhang0J0Z024} & 27.125 & 0.082 & 0.152 & 0.684 & 0.427 & 0.957 & 0.281 & 0.892 & 0.946 & 0.928 & 0.05 & 19:15 \\
        RAVE \cite{DBLP:conf/cvpr/KaraKYRY24} & 26.318 & 0.094 & 0.138 & 0.742 & 0.374 & 0.960 & 0.286 & 0.915 & 0.961 & 0.947 & 0.08 & 20:40 \\
        Slicedit \cite{DBLP:conf/icml/CohenKKHM24} & 25.692 & 0.112 & 0.148 & 0.715 & 0.389 & 0.958 & 0.280 & 0.908 & 0.953 & 0.940 & 0.05 & 28:11 \\
        VideoGrain \cite{DBLP:conf/iclr/YangZF025} & 28.416 & 0.058 & 0.113 & \textbf{0.768} & 0.331 & \textbf{0.971} & 0.284 & 0.927 & 0.968 & 0.957 & 0.12 & 13:30\\
        VidToMe + ControlNet-Depth \citep{DBLP:conf/cvpr/LiMYY24} & 26.683 & 0.121 & 0.165 & 0.598 & 0.491 & 0.958 & 0.273 & 0.892 & 0.946 & 0.920 & 0.05 & 12:37 \\
        VidToMe + PnP \citep{DBLP:conf/cvpr/LiMYY24} & 27.820 & 0.065 & 0.114 & 0.706 & 0.403 & 0.962 & 0.293 & 0.915 & 0.960 & 0.940 & 0.08 & 15:26 \\        
        TSPVE + ControlNet-Depth & 27.915 & 0.096 & 0.188 & 0.697 & 0.483 & 0.956 & 0.271 & 0.912 & 0.950 & 0.938 & 0.12 & \textbf{10:40} \\
        TSPVE + PnP & \textbf{28.629} & \textbf{0.051} & \textbf{0.111} & 0.774 & \textbf{0.326} & 0.969 & \textbf{0.295} & \textbf{0.932} & \textbf{0.972} & \textbf{0.962} & \textbf{0.20} & 13:24 \\
        \bottomrule
    \end{tabular}
     }
    \caption{Quantitative evaluation results.}\label{tab:quantitative_comparison}  
\end{table*}

Table \ref{tab:quantitative_comparison} shows the quantitative evaluation, covering comparisons with both one-shot and zero-shot editing methods.
In terms of temporal consistency, our method integrated with the controlling scheme of Plug and Play (PnP) performs the best, outperforming all compared baselines and SOTA methods, and verifying its superiority in maintaining inter-frame temporal smoothness.
To quantify our core target of source video temporal structure preservation, we adopt the proposed Temporal Structure Preservation Rate (TSPR) metric. Results show our method achieves optimal performance on temporal structure-related metrics, providing direct evidence for the effectiveness of our core designs, and addressing the key limitation of existing methods that overlook high-level semantic temporal structure.
In regard to text alignment and user voting, our approach also gives competitive performance, with edited results more consistent with human visual preference.
Apart from the last four rows of token merging based methods, to edit long videos (in this experiment, we run videos consisting of 240 frames generally) and avoid running out of GPU memory, other methods have to cut the video into several sub-videos and edit them individually and recombine finally, which even damages the video’s global temporal structure and introduces extra overhead.
The editing time per video on average is shown in the last column. Tune-A-Video\citep{DBLP:conf/iccv/WuGWLGSHSQS23} and Video-P2P\citep{DBLP:conf/cvpr/LiuZ00J24} not only present poor temporal consistency of edited video but also need to take much more time (over an order of magnitude) due to video-specific fine-tuning. AdaFlow\citep{DBLP:journals/corr/abs-2502-05433} achieves relatively efficient inference via attention optimization, while ours maintains acceptable running time while preserving temporal structural consistency, achieving a well-balanced trade-off between editing effectiveness and computational efficiency.

\subsection{Ablation Studies}

\begin{figure}[t]
    \centering
        \includegraphics[width=1.0\linewidth]{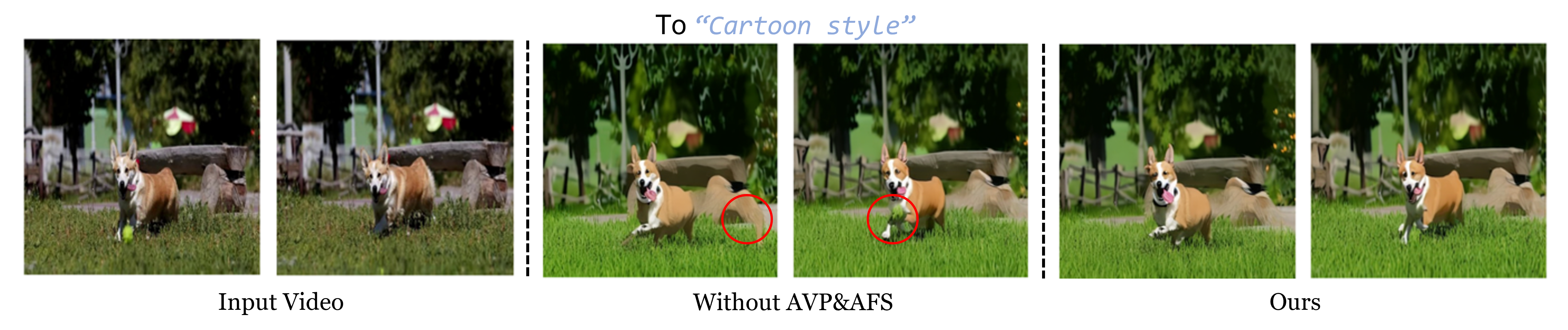}
        \caption{Ablation on Adaptive Video Partition and Anchor Frames Selection (AVP-AFS). Without capturing the source video's temporal structure using AVP-AFS, object misalignment and abrupt changes often occur.}
        \label{fig:Ablation_1}  
\end{figure}

\subsubsection{Adaptive Video Partition and Anchor Frames Selection}
We conduct an ablation experiment on the scheme of Adaptive Video Partition and Anchor Frames Selection (AVP-AFS) as shown in the Figure
\ref{fig:Ablation_1}. 
We capture the temporal structure of the source video using AVP-AFS and impose its influence on the denoising diffusion process. Without this operation, the method cannot well keep the same temporal structure between the edited video and the original one, easily causing object misalignment and abrupt changes, e.g., a dog's paw parts disappear or hind legs suddenly appear in the background.

To further verify the rationality of our Anchor Frame Selection (AFS) design, we conduct comparison experiments comparing our strategy with two alternative schemes: (1) selecting the temporal midpoint of the clip; (2) random frame sampling within the clip. Our strategy selects the frame with the maximum row sum of the intra-clip cosine similarity matrix as the anchor, which corresponds to the semantic centroid of the clip and has reasonable clip representativeness.

\begin{table}[h]
\centering
\footnotesize
\renewcommand\arraystretch{1.1}
\setlength{\tabcolsep}{6pt}
\begin{tabular}{lcc}
\toprule
\textbf{Strategy} & \textbf{TSPR $\uparrow$} & \textbf{Inp. Err. $\downarrow$} \\
\midrule
Temporal Midpoint  & 0.887 & 0.127 \\
Random Frame Sampling  & 0.862 & 0.143 \\
AFS (Ours)  & \textbf{0.962} & \textbf{0.051} \\
\bottomrule
\end{tabular}
\caption{Comparison of Anchor Frame Selection Strategies}
\label{tab:anchor_compare}
\end{table}
The results in Table \ref{tab:anchor_compare} show that our strategy achieves the best performance on both metrics, significantly outperforming the two alternatives. This validates that our anchor frame selection strategy can provide a semantically representative reference substance for subsequent token merging, effectively improving the temporal structure preservation performance of the edited video.

\begin{figure}[t]
    \centering
     \includegraphics[width=1.0\linewidth]{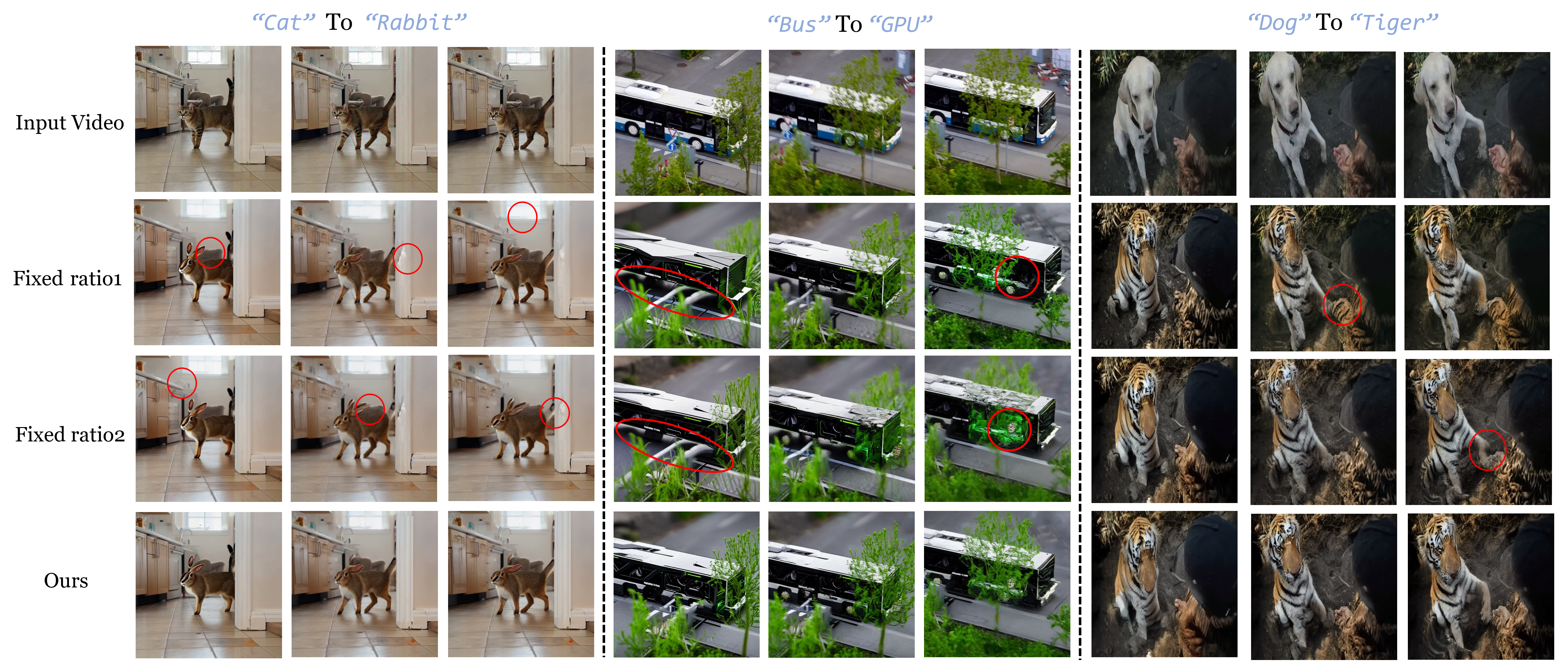}
    \caption{Ablation on Clip-adaptive Token Merging: At a low merging ratio (Fixed ratio1=0.5), there are significant discrepancies in the sofa's shape, door frame and window grid between different clips. At a high fixed merging ratio (Fixed ratio2=0.9), the running tap vanishes, and the parts of the rabbit's back or tail blend with the background.}
    \label{fig:Ablation_2}  
\end{figure}
\paragraph{Clip-Adaptive Token Merging}
We ablate the clip-adaptive token merging module by setting fixed merging ratios for all clips. We experiment with two different fixed ratios ($p$ = 0.9 and 0.5). The experimental results are shown in Figure \ref{fig:Ablation_2}. Clearly, using fixed ratios can lead to improper merging, causing inconsistencies in similar regions of the video. In contrast, our clip-adaptive merging strategy, which utilizes spatiotemporal window-based similarity measurements and L1 distance to filter redundant tokens, achieves more consistent and semantically coherent results, significantly outperforming the fixed ratio approach.

\paragraph{Alternated Combination of Adjacent Clips}
We also ablate the alternated combination strategy for cross-clip joint self-attention to compare our method with two reduced settings: (i) No cross-clip joint self-attention, in which self-attention is performed only for individual clips without any cross-clip interaction; (ii) Fixed clip combination layout, in which adjacent clips are combined in one fixed arrangement (selected from our odd or even timestep arrangements) for joint self-attention without alternation. As shown in Table \ref{tab:Ablation_1}, the comparisons w.r.t two metrics show the necessity of our alternation strategy in enhancing inter-clip continuity. 

\begin{table*}[t]
    \centering
    
    \resizebox{\textwidth}{!}{ 
    \begin{tabular}{lccccccc}  
        \toprule
        & \multicolumn{5}{c}{Temporal Consistency} & \multicolumn{2}{c}{Text Alignment} \\
        \cmidrule(lr){2-6} \cmidrule(lr){7-8}  
        & PSNR$\uparrow$ & Inp. Err.$\downarrow$ & Warp Error$\downarrow$ & SSIM $\uparrow$ & LPIPS $\downarrow$ & Frame CLIP Score$\uparrow$ & Text CLIP Score$\uparrow$ \\  
        \midrule
        No inter-clip & 25.517 & 0.127 & 0.176 & 0.755 & 0.424 &   0.947 & 0.270  \\ 
        Fixed neighbor & 27.193 & 0.104 & 0.122 & 0.748 & 0.374 & 0.953 & 0.277  \\
        TSPVE  & \textbf{28.629} & \textbf{0.051} & \textbf{0.111} & \textbf{0.774} & \textbf{0.326} & \textbf{0.969} & \textbf{0.284} \\
        \bottomrule
    \end{tabular}
    }
    \caption{Ablation on Alternated Combination Strategy}\label{tab:Ablation_1}  
\end{table*}